\documentclass[conference]{IEEEtran}
\IEEEoverridecommandlockouts
\usepackage{cite}
\usepackage{amsmath,amssymb,amsfonts}
\usepackage{algorithmic}
\usepackage{graphicx}
\usepackage{textcomp}
\usepackage{xcolor}
\def\BibTeX{{\rm B\kern-.05em{\sc i\kern-.025em b}\kern-.08em
    T\kern-.1667em\lower.7ex\hbox{E}\kern-.125emX}}
\begin{document}

\title{Agentic Pipeline for Self-Synchronized Multiview Joint Angle Monitoring in Uncalibrated Environments}
\author{Juncheng Yu$^{1}$, Lusi A$^{1}$, Haoxuan Xie$^{1}$, and Weiming Wang$^{1}$$^\star$
\thanks{$^{1}$ Juncheng Yu, Lusi A, Haoxuan Xie, and WeimingWang are with the  National Engineering Research Center of Neuromodulation, School of Aerospace Engineering, Tsinghua University, Beijing 100084, China.}
\thanks{$^\ast$ Corresponding author.}
\thanks{© 2026 IEEE. Personal use of this material is permitted. Permission from IEEE must be obtained for all other uses, in any current or future media, including reprinting/republishing this material for advertising or promotional purposes, creating new collective works, for resale or redistribution to servers or lists, or reuse of any copyrighted component of this work in other works. This paper has been accepted by IEEE EMBC 2026. The final published version will be available on IEEE Xplore.}%
}

\maketitle

\begin{abstract}
    Kinematic monitoring plays a critical role in long-term rehabilitation for patients with spinal cord injury (SCI), where multi-view markerless motion capture methods have shown significant potential. However, owing to the reliance on calibration and the difficulty of achieving multi-view synchronization, their deployment in patient self-deployed environments remains challenging. In this work, we propose an agentic pipeline for self-synchronized multi-view joint angle monitoring in uncalibrated environments using two cameras without hardware triggers. The Multimodal large language models enable automatic video synchronization and agent-driven self-verification. State-of-the-art monocular 2D pose estimation models are employed to extract candidate poses, where an agent-based selection mechanism is then applied to automatically identify and track the target subject, thereby producing consistent 2D poses in the presence of multiple individuals and occlusions. Such 2D poses are optimized to estimate joint angles from uncalibrated multi-view pose sequences, ensuring interpretability through explicit geometric modeling. Validation against Vicon system demonstrated the strong performance, achieving an MAE of $5.97^\circ \pm 2.36^\circ$ and a Pearson correlation coefficient of $0.962 \pm 0.014$. The proposed method is expected to provide a practical, patient self-deployable system to perform daily kinematic monitoring in uncalibrated home environments.
\end{abstract}

\section{Introduction}
Daily kinematic monitoring plays a critical role in long-term neurorehabilitation\cite{lee2024wearable}\cite{dobkin2013wearable} for patients undergoing spinal cord stimulation (SCS)\cite{hankov2025augmenting}, which provides quantitative and objective metrics to assess motor function and recovery over time.
However, traditional kinematic analysis methods utilizing sensors such as motion capture systems \cite{lili2021quantifying}\cite{bellitto2023clinical} or wearable devices \cite{wolf2024decoding}\cite{kempske2023identifying} are often limited by factors such as space, time, and the need for marker placement, restricting their use in home, clinic, and rehabilitation rooms \cite{cotton2020kinematic}, and in many cases, the patients need to stand or even squat for sensor calibration, which is particularly challenging for SCI patients.
Therefore, there is a need for a kinematic analysis method that reduces the burden on SCI patients during rehabilitation so that the daily monitoring can be conducted more conveniently and effectively.

Markerless methods based on computer vision have been proposed to address the limitations \cite{d2021validation}\cite{gu2018markerless}\cite{yunardi2023motion}.
Two-dimensional (2D) markerless methods have been most commonly used \cite{wang2021single}\cite{menychtas2023gait}, as the 2D methods are easy to be implemented and require only a single RGB camera.
However, 2D-based kinematic analysis inherently lacks depth information, which limits its ability to capture consistent joint motion under varying viewpoints and subject postures, leading to kinematically inconsistent measurements across time and sessions.
Such limitations significantly reduce the reliability of 2D kinematic metrics in unconstrained clinical and home-based rehabilitation scenarios.

Depth estimation methods that lift the pixel coordinates to 3D coordinates are therefore proposed, which are also known as three-dimensional (3D) markerless methods.
Two main approaches are used to get 3D information: monocular depth estimation via neural networks \cite{shimada2021neural}\cite{shi2020motionet} and stereo vision \cite{d2021validation}\cite{slembrouck2020multiview}.
Monocular depth estimation methods based on a single RGB camera offer a convenient solution for generating 3D information, but the reliance on neural networks limits their interpretability, making it difficult for clinicians and researchers to perform reliable long-term comparative analysis of kinematic data and thereby constraining their applications in clinical and research settings.
In contrast, stereo vision methods can provide mathematical and traceable results with high-confidence, enabling consistent and comparable kinematic measurements across time and sessions.
However, stereo vision methods typically require precise intrinsic and extrinsic camera calibration, rigid multi-camera setups, and accurate inter-camera synchronization, which limit their practicality in home-based rehabilitation scenarios, where patients are expected to deploy the system using general devices such as smartphones, making camera calibration and synchronized camera setup largely impractical.

In this paper, we propose an agentic pipeline for self-synchronized multiview joint angle monitoring in uncalibrated environments, specifically designed for long-term, home-based rehabilitation scenarios.
The proposed pipeline can provide joint angles from estimated 3D poses, which requires only two consumer-grade RGB cameras, such as smartphones, without relying on pre-calibrated intrinsic or extrinsic parameters, nor on hardware-level synchronization.
Multimodal large language models (MLLMs) with visual capabilities are integrated into the pipeline as autonomous agents to guide synchronization, verification, and quality control processes, enabling reliable and interpretable kinematic measurements without manual intervention.
With multi-view video sequences synchronized by the MLLMs, state-of-the-art monocular 2D pose estimation models are applied independently to each view to extract candidate 2D poses, which are then iteratively selected by the agents to ensure temporally and cross-view consistent target poses.
To prevent the exposure of identifiable information, facial anonymization is performed locally prior to online MLLM processing by leveraging lightweight face detection and blurring techniques.
Reliable joint angles are then estimated using optimization-based 2D–3D lifting with explicit geometric constraints, facilitating cross-session comparability and improving interpretability.
We also conducted synchronous experiments with Vicon motion capture system to validate our method, and evaluated the accuracy and reliability of the proposed method in joint angle tracking.
Promising results were achieved by visualizing the angles and quantitatively evaluating the MAE and Pearson correlation coefficients from both methods, demonstrating the reliability of the proposed method in the typical scenarios and applications.

To summarize, the main contributions of this paper are as follows:
\begin{itemize}
  \item We propose an agentic pipeline for self-synchronized multiview joint angle monitoring in uncalibrated environments, enabling practical and patient self-deployable kinematic analysis for long-term, home-based rehabilitation without requiring camera calibration or hardware-level synchronization.

  \item Multimodal large language models (MLLMs) are proposed to serve as agentic controllers to enable autonomous multi-view synchronization, target pose selection, and iterative quality control, thereby supporting patient self-deployable kinematic data acquisition with common devices in uncalibrated home environments.

  \item Synchronous experiments are conducted to compare results with the gold standard from Vicon. The proposed method achieves an MAE of $5.97^\circ \pm 2.36^\circ$ and a Pearson correlation coefficient of $0.962 \pm 0.014$, demonstrating its reliability and accuracy.
\end{itemize}

\section{Methods}
\begin{figure*}[!t]
    \vspace{2mm} 
    \centering
    \includegraphics[width=0.8\linewidth]{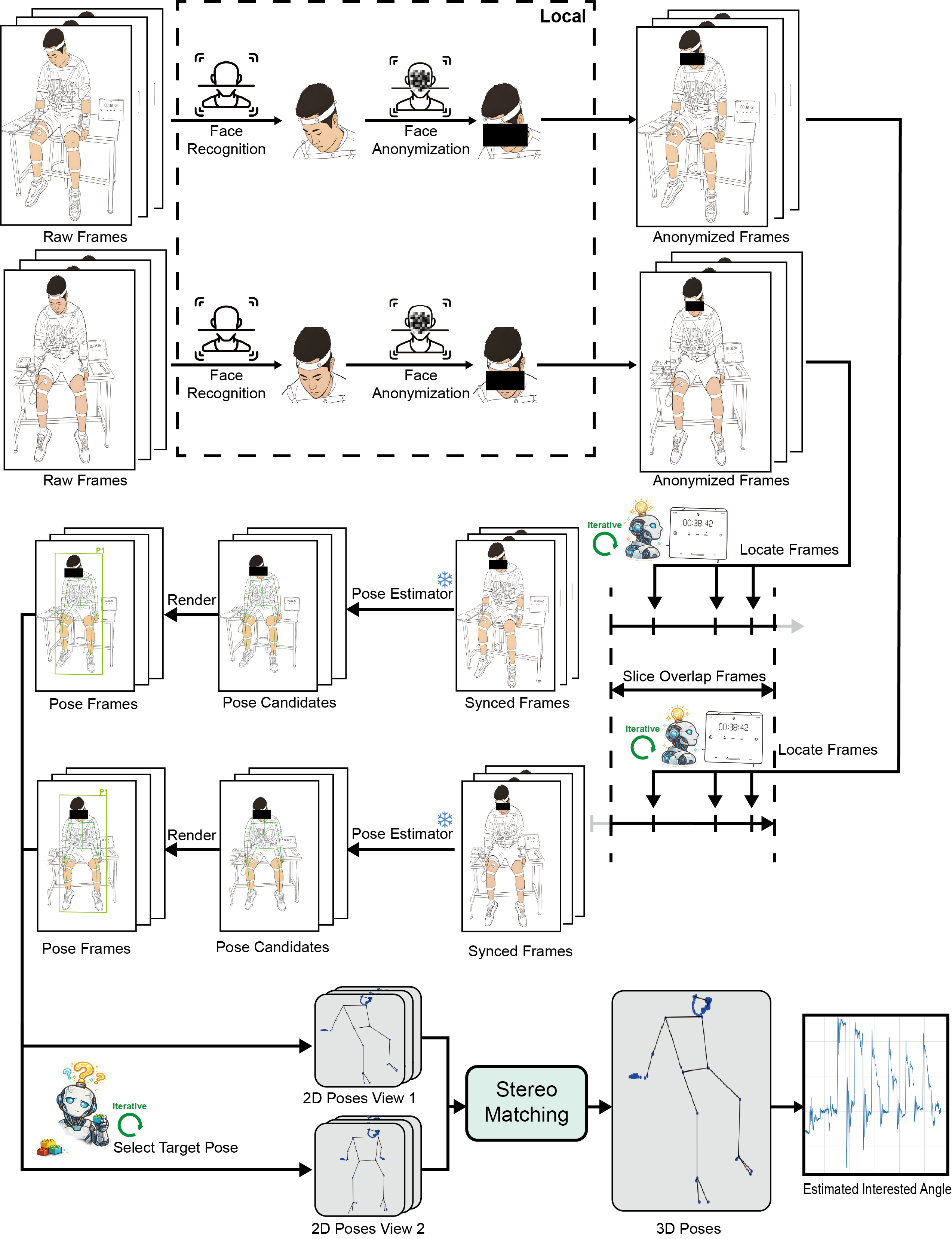}
    \caption{Overview of the proposed method. 
    The input RGB frames are first fed into face recognition and anonymization modules to generate privacy-preserving frames. 
    The anonymized frames are then analyzed by MLLM–driven agents, which iteratively infer frame-level temporal observations for multi-view synchronization. 
    The synchronized multi-view frames are subsequently used for monocular 2D pose estimation independently in each view. 
    All estimated poses within each frame are rendered with bounding boxes and identifiers, enabling the MLLM-driven agents to identify the primary target and obtain continuous target 2D pose sequences across views. 
    Such selected 2D joint coordinate sequences from two cameras are then fed into the stereo lifting module to estimate 3D joint coordinates.
    Finally, the estimated 3D pose are used to compute the target joint angles.}
    \label{fig:overview}
\end{figure*}

An overview of the proposed method is shown in Fig. \ref{fig:overview}. 
Given asynchronous multi-view RGB videos captured by two uncalibrated consumer devices, multimodal agents autonomously reason to establish synchronization, maintain identity consistency, and enable reliable joint angle estimation.
The input frames are first anonymized locally to preserve privacy, after which MLLM-driven agents operate on the anonymized frames to extract frame-level temporal observations from shared visual clocks and infer cross-view temporal correspondence, enabling reliable multi-view synchronization without hardware support.
The synchronized view is processed independently using a monocular 2D human pose estimation model to obtain candidate multi-person pose predictions, where all detected poses are retained and rendered with indexed visual annotations, allowing the MLLM-driven agents to identify the primary target subject and perform iterative quality control.
Finally, the synchronized and identity-consistent 2D joint trajectories from the two views are fed into a geometry-based stereo lifting module.
Without relying on pre-calibrated camera parameters, the system estimates epipolar geometry and reconstructs 3D joint trajectories through triangulation and bundle adjustment.
The resulting 3D poses are then used to compute joint angles of interest, enabling cross-session consistent and interpretable joint angle tracking for further kinematic analysis.
\subsection{Agentic Synchronization of Asynchronous Multi-view Videos}
In home-based rehabilitation scenarios, multi-view videos are typically captured using consumer-grade devices such as smartphones or tablets, where hardware-level synchronization is unavailable and device timestamps are often unreliable owing to clock drift and frame rate variability.
To address that, a MLLM-driven agent is utilized to identify the shared visual clock such as an IPAD screen in the video frames and infer temporal correspondence across asynchronous video streams.
Given a frame containing a visible clock display, the agent is instructed to determine whether a tablet device is present and assesses its visibility, attempts to read any timestamp displayed on the device screen, infers relative temporal information without estimating absolute real-world time, and arranges the extracted temporal observations into a structured format.

\textit{Example: You are a strict image analysis assistant. For each input image, you are given the corresponding video filename and frame index. Your task is to identify whether a tablet screen (e.g., an iPad) is present and to read any valid timestamp visible in the frame (preferably from the tablet display). Record the timestamp exactly as observed without guessing or completion; if no valid timestamp is visible, return null. Output only a valid JSON object containing the video filename, frame index, detection status, the extracted timestamp value (or null), and a brief note explaining the decision.}

However, querying the MLLM for timestamp extraction on every video frame is computationally prohibitive and impractical.
With redundant visual information across consecutive frames, we combine the MLLM-based timestamp extraction with a temporal propagation strategy to balance efficiency and accuracy.
The timestamp extraction module employs a hierarchical adaptive sampling scheme to reduce the number of expensive vision-based API calls and ensure accurately detect temporal drift over long video sequences. 
The candidate timestamps are first extracted from a small set of adaptively sampled frames distributed across the entire video duration, which are all processed by the MLLM-based timestamp extraction agent.
The extracted timestamps are then analyzed to estimate temporal drift with respect to the nominal frame rate, allowing the identification of temporally stable regions within the video, which serves as a reference for following temporal propagation.
Timestamps for the remaining frames are then propagated from the sampled observations, avoiding redundant MLLM queries while preserving temporal consistency.
To prevent error accumulation caused by frame duplication and frame rate variability, additional refinement samples are adaptively selected beyond the stable region to verify drift consistency and correct potential deviations.
Finally, a secondary validation stage is conducted using a small set of verification samples drawn from stable regions, temporal boundaries, and randomly selected frames, which ensure the synchronization quality and detect unexpected anomalies that may arise from visual ambiguity or frame irregularities.

Besides, to facilitate MLLM-based recognition, each sampled frame is preprocessed with gray-world white balancing and contrast-limited adaptive histogram equalization to enhance the readability of white timestamp text.
Notably, the MLLM operates directly on the entire video frame rather than predefined cropped regions, whose robust multimodal visual understanding capability allows reliable identification and interpretation of timestamp text within complex scenes without explicit localization prior.
To avoid the exposure of identifiable information to API providers, facial anonymization is performed locally prior to MLLM processing.
A lightweight face detection model is applied to identify facial regions in each submitted frame, which are anonymized through blurring before being forwarded to the MLLM.
\subsection{Target Human Pose Estimation in 2D}
The 2D human pose estimation always utilizes two-stage networks, including top-down and bottom-up methods.
The top-down methods first detect the human bounding box and then estimate the 2D joint coordinates within the bounding box while bottom-up methods first detect the 2D joint coordinates and then group them into human poses. 
In this paper, we adopted a pretrained top-down human pose estimation model Sapiens \cite{khirodkar2024sapiens} to estimate the 2D joint coordinates, which was a vision foundation model that provides a comprehensive suite for human-centric tasks.

In the top-down human pose estimation framework, given input RGB image $I \in \mathbb{R}^{H \times W \times 3}$ with height $H$, width $W$, and three channels, the video can be represented as a sequence of images $V = \{I_1, I_2, ..., I_T\}$.
Each video from two cameras is treated as a sequence of individual images, and processed independently in the pose estimation framework, which means the estimated 2D joint coordinates are obtained for each frame from two cameras respectively.

In practice, the top-down pose estimator outputs 2D joint predictions for all detected human instances in each frame.
Rather than selecting a single subject at this stage, we retain all raw pose estimation results as $\mathbf{P}_{\mathrm{raw}} \in \mathbb{R}^{T \times N \times J \times 3}$, to enable subsequent target subject identification and quality control,
where $T$ denotes the total number of frames in the video, $J$ is the number of body joints defined by the pose model, the last dimension corresponds to the $(x, y)$ image coordinates and the associated confidence score for each joint, and $N$ represents the maximum number of detected human instances across the entire video sequence, follows
\begin{equation}
N = \max_{t \in \{1,\dots,T\}} N_t ,
\end{equation}
where $N_t$ is the number of detected persons in frame $t$.
For frames containing fewer than $N$ detected persons, empty entries are padded accordingly. 

To select the target subject from the retained multi-person pose set, we introduce an agent-driven identity identification and quality control module designed for unconstrained multi-person scenarios, which combined sparse visual recognition with trajectory-based propagation and validation.

Specifically, a small set of representative frames is adaptively sampled from the video sequence in terms of multi-person trajectory proximity and temporal instability.
For each sampled frame, all detected human poses are rendered with indexed skeleton overlays and bounding boxes and provided to a multimodal vision-language model, where the primary target subject is determined based on the visual frequency of appearance across multiple sampled frames.
The agent-identified target indices from sampled frames serve as anchor points for identity propagation, which utilizes a Kalman filter–based tracking scheme to propagate identity assignments across adjacent frames.

Specifically, for each anchor frame, we initialize a 2D Kalman filter whose state corresponds to the target’s image-plane center computed as the confidence-weighted centroid of all detected joints, follows
\begin{equation}
    \mathbf{c}_{t}^{(n)} = \frac{\sum_{j=1}^{J} c_{t,j}^{(n)} \cdot \begin{pmatrix} x_{t,j}^{(n)} \\ y_{t,j}^{(n)} \end{pmatrix}}{\sum_{j=1}^{J} c_{t,j}^{(n)}},
\end{equation}
The filter is warmed up using the most recent consecutive non-empty target assignments to obtain a stable estimate of the target’s location and uncertainty, during which the filter is recursively updated with the corresponding weighted centers, allowing the state estimate to converge before propagation.
For each subsequent frame $t$, the Kalman filter predicts the target center $\hat{\mathbf{c}}_t$, and the identity is assigned by selecting, among all valid person candidates, the instance whose weighted center $\mathbf{c}_{t}^{(n)}$ is closest to $\hat{\mathbf{c}}_t$ in Euclidean distance.
Association is then performed by selecting, among all valid person candidates in frame $t$, the instance whose weighted center $\mathbf{c}_{t}^{(n)}$ is closest to the predicted target center $\hat{\mathbf{c}}_t$ in Euclidean distance:
\begin{equation}
n_t^* = \arg\min_{n \in \mathcal{V}_t} \left\| \mathbf{c}_{t}^{(n)} - \hat{\mathbf{c}}_t \right\|_2 ,
\end{equation}
where $\mathcal{V}_t$ denotes the set of valid person candidates at frame $t$.
The selected candidate is then treated as the measurement to update the Kalman filter, producing a temporally smoothed estimate of the target trajectory.
During the Kalman-based propagation, the agent-identified anchor frames are explicitly used to correct the filter state and prevent drift accumulation.
Specifically, when a frame corresponds to an anchor identified by the vision-language model, the Kalman filter update is conditioned directly on the anchor-associated weighted center rather than relying on the predicted state propagated from previous frames.

To suppress identity switches in crowded scenes, a bounding-box consistency constraint is adopted during propagation.
For each candidate pose, a tight bounding box is computed from its valid joints, and the intersection-over-union (IoU) with the most recent accepted target bounding box is evaluated.
The candidate will be rejected and the frame is marked as missing if the IoU is almost zero, while the last accepted bounding box is retained for subsequent matching, which prevents frames without a reliable target pose from adversely affecting subsequent identity propagation and preserves temporal continuity.
Clean and temporally consistent target-specific 2D pose sequences are therefore obtained for each view, enabling further multiview triangulation and kinematic analysis.
\begin{figure*}[!t]
    \centering
    \vspace{2mm} 
    \includegraphics[width=0.9\linewidth]{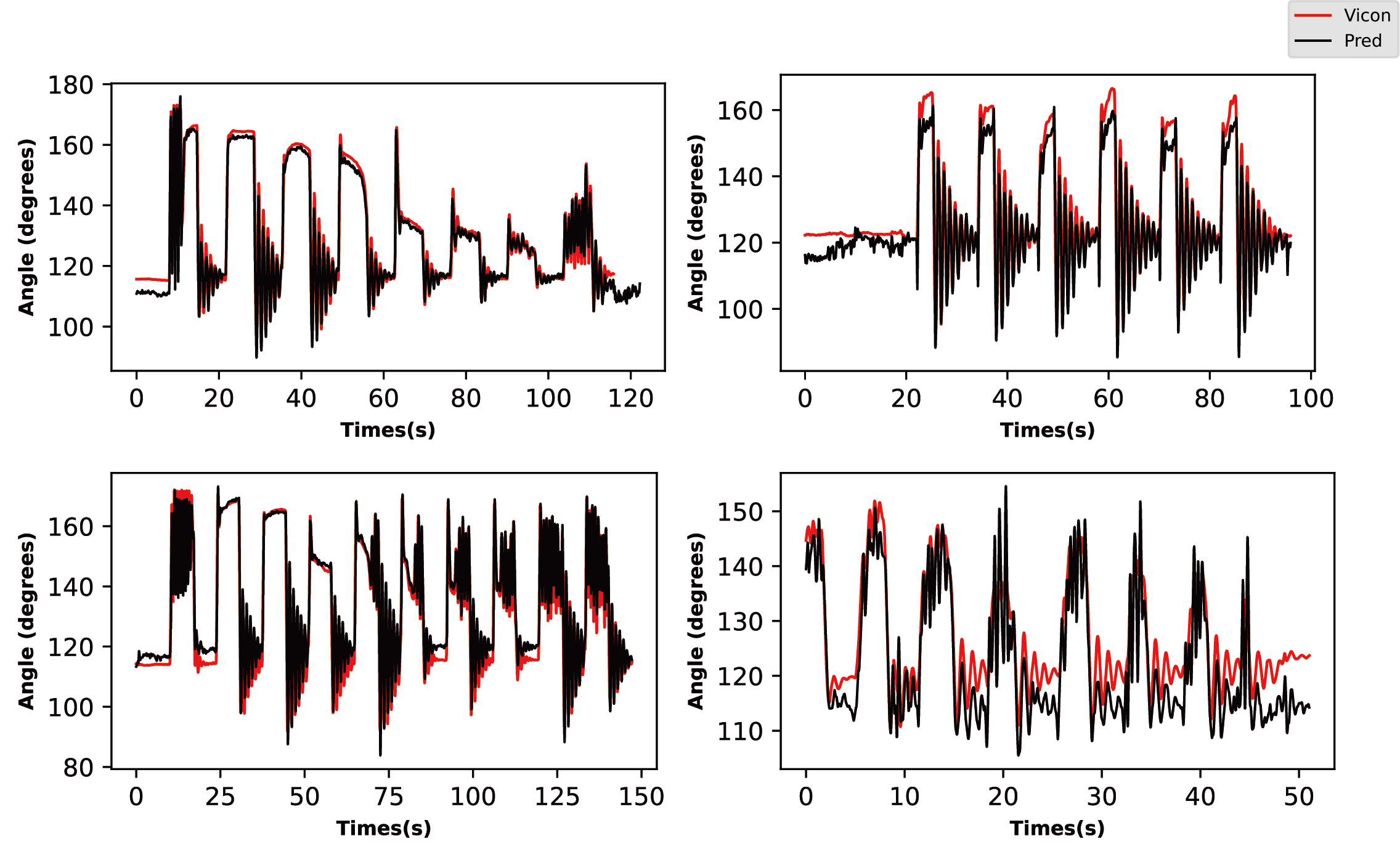}
    \caption{The joint angles from the proposed method and the golden standard from the Vicon system. The black line is the estimated joint angles from the proposed method and the red line is the golden standard from the Vicon system.}
    \label{fig:results}
\end{figure*}
\subsection{Lifting 2D to 3D Mathematically}
Estimating 3D joint coordinates from two synchronized 2D pose sequences requires establishing geometric correspondence between the two camera views. 
In calibrated stereo settings, this is typically achieved using known intrinsics $(\mathbf{K}_1,\mathbf{K}_2)$ and extrinsics $[\mathbf{R}\,|\,\mathbf{t}]$.
However, in self-deployed home scenario, neither reliable intrinsic nor extrinsic calibration is available, therefore the camera geometry must be inferred from the observed 2D joint trajectories.

Given the identity-consistent 2D pose sequences $\mathbf{P}_1=\{ \mathbf{u}_{1,t}\}_{t=1}^{T}$ and $\mathbf{P}_2=\{ \mathbf{u}_{2,t}\}_{t=1}^{T}$ obtained from two views, we first estimate the epipolar geometry by fitting a fundamental matrix $\mathbf{F}$ using a RANSAC-based estimator on high-confidence joint correspondences aggregated across frames, where an inlier threshold $\tau_F$ defined in the normalized coordinate system is used to determine the consensus set and reject outliers.

Furthermore, to handle unknown intrinsic parameters, we initialize pseudo-intrinsics $\mathbf{K}_1^0$ and $\mathbf{K}_2^0$ using an image-centered pinhole model with focal length proportional to the image size, where an initial essential matrix is then formed as $\mathbf{E}_0 = \mathbf{K}_2^{0\top}\mathbf{F}\mathbf{K}_1^0$.

With $(\mathbf{R}_0,\mathbf{t}_0)$ and $(\mathbf{K}_1^0,\mathbf{K}_2^0)$, we triangulate all joints across frames to obtain an initial 3D reconstruction.
To further improve robustness and kinematic consistency without relying on any pre-calibrated parameters, we refine the 3D joint trajectories via an optimization-based bundle adjustment that minimizes the multi-view reprojection error while incorporating geometric and kinematic regularization.
Specifically, we refine the initial triangulated 3D joints by solving a bundle-adjustment problem that jointly optimizes the 3D joint coordinates $\mathbf{X}$ and the relative pose $(\mathbf{R},\mathbf{t})$, while keeping the pseudo-intrinsic matrices $(\mathbf{K}_1^0,\mathbf{K}_2^0)$ fixed.
The optimization is performed using the Adam optimizer and minimizes a composite objective consisting of a reprojection error and an epipolar consistency term based on the Sampson distance induced by the estimated fundamental matrix.
The refined 3D joint trajectories are then used for downstream joint angle estimation.
\section{Experiments and Results}
\subsection{Implement Details}
In this paper, we employ the GLM-4.5V multimodal vision-language model as foundation model of the agent, to recognize millisecond-level timestamps displayed on an iPad screen within video frames, and identify the primary target subject.
The RTMPose detector \cite{RTMPose} implemented in mmdetection \cite{mmdetection} is first used to detect the human bounding box in the image, which will be used as the input of the top-down human pose estimation model.
With such bboxes, we estimate 2D pose sequence from each camera by utilizing Sapiens with 2.0B parameters and get whole body 2D joint coordinates with totally 133 joints, following coco whole body keypoints format.
The threshold for selected points is set to 0.98, where we find higher threshold will lose dynamic points we interested in and lower threshold will introduce more noise.
The golden standard was achieved by Vicon system and synchronized manually with visual trigger, that is a timer with precision up to three decimal places in milliseconds.
The settings of the Vicon system are 39 keypoints plug-in gait model, 100Hz sampling rate.
The experimental dataset consists of a healthy subject and two patients with SCI, where the SCI patients were recorded at multiple follow-up sessions, whereas the healthy subjects performed the experimental protocols once and served as reference measurements for comparison.

\subsection{Results}
We first evaluated the temporal alignment accuracy achieved by the proposed MLLM-driven multi-view synchronization with propagation.
Using the visually inspected timestamps from the shared visual clock as ground truth, we evaluated the temporal alignment by comparing the theoretical frame indices corresponding to the same timestamp across views for each synchronized video pair.
The agent-based synchronization achieved a mean absolute temporal error of 32.4 ms with a standard deviation of 12.1 ms, demonstrating reliable cross-view alignment without hardware synchronization.

\begin{figure}[!htbp]
    \centering
    \vspace{2mm} 
    \includegraphics[width=0.9\linewidth]{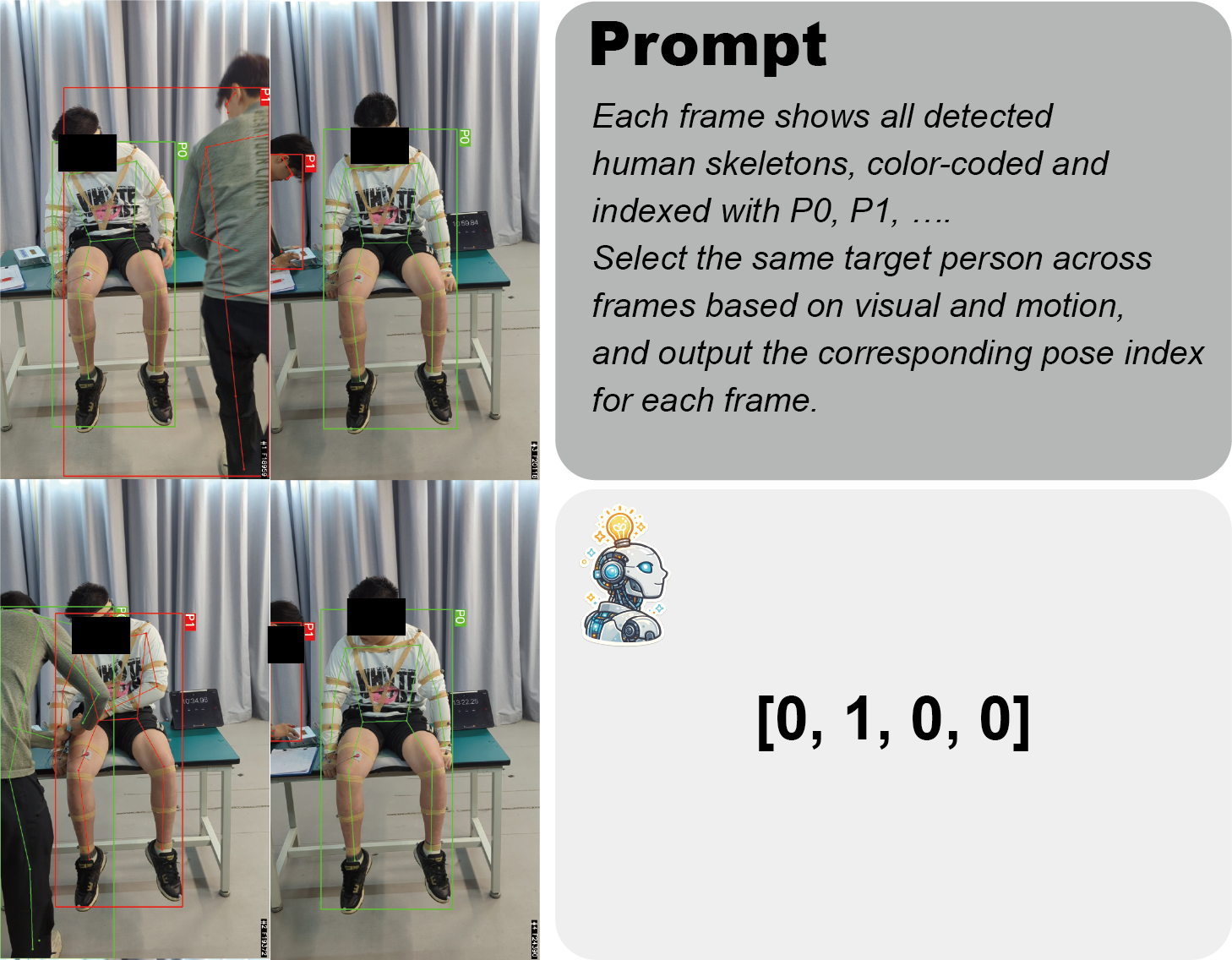}
    \caption{Representative example of the input and output of the MLLM-driven target identification module.}
    \label{fig:results_indentify}
\end{figure}

To illustrate the behavior of the proposed MLLM-driven target identification module, we provide an example of its input–output mapping on a long video sequence with 30 minutes.
For each query, the input consists of multiple frames in which all detected human poses are rendered with indexed skeletons such as P0, P1, allowing the MLLM to infer the primary target based on visual consistency across frames, and The output is the index of the pose corresponding to the primary target subject in each frame.
Both the predicted and manually annotated labels take values in the range $[-1, N_t-1]$, where $-1$ indicates that no valid target pose is present in the frame, and a non-negative value denotes the index of the selected target pose among all detected poses in that frame.
Fig.~\ref{fig:results_indentify} visualizes a representative example of the input and output of the MLLM-driven target identification module.

We validated the joint angles using two metrics: the MAE,
and the Pearson correlation coefficient of the joint angles to assess the overall consistency between the measured values and the true angles.

The MAE is calculated by Eq. \ref{eq:MAE}.
\begin{equation}
    MAE = \frac{1}{N} \sum_{i=1}^{N} |\theta_{est}^{(i)} - \theta_{gt}^{(i)}|
    \label{eq:MAE}
\end{equation}
where $\theta_{est}$ is the estimated joint angles from the proposed method, $\theta_{gt}$ is the golden standard from the Vicon system, and $N$ is the number of frames in the video.
We achieved an MAE of 5.97 $\pm$ 2.36 among angles ranging from 0 to 180 degrees.

The pearson correlation coefficient is calculated by Eq. \ref{eq:pearson}.
\begin{equation}
    r = \frac{\sum_{i=1}^{N} (\theta_{est}^{(i)} - \bar{\theta}_{est})(\theta_{gt}^{(i)} - \bar{\theta}_{gt})}{\sqrt{\sum_{i=1}^{N} (\theta_{est}^{(i)} - \bar{\theta}_{est})^2} \sqrt{\sum_{i=1}^{N} (\theta_{gt}^{(i)} - \bar{\theta}_{gt})^2}}
    \label{eq:pearson}
\end{equation}
where $\bar{\theta}_{est}$ and $\bar{\theta}_{gt}$ are the mean values of the estimated joint angles and the golden standard joint angles, respectively.
We achieved pearson correlation coefficients across all measurements achieved 0.962 $\pm$ 0.014.

We also compared the joint angles to provide a clear visualization of our method’s ability to capture fine-grained variations in joint angles, which are calculated by the minimum and maximum values of the joint angles in the whole sequence during one measurement.
The results are shown in Fig. \ref{fig:results}.

As shown in Fig. \ref{fig:results}, the estimated joint angles from the proposed method are highly consistent with the golden standard from the Vicon system.
The proposed method performs well in clean data from normal subjects, following the golden standard closely.
Moreover, the proposed method also performs well in noisy data from SCI patients, which also capture the fine-grained swing in SCI patients' joint angles.
The promising results show the high accuracy and reliability of the proposed method in joint angle tracking.

\section{Conclusion}
In this paper, we proposed a self-synchronized, mathematically grounded markerless pipeline for joint angle tracking in spinal cord injury (SCI) rehabilitation scenarios.
The method estimates joint angles from two uncalibrated RGB cameras without relying on hardware-level synchronization or pre-calibrated camera parameters, addressing practical constraints in home-based settings.
By integrating multimodal large language models (MLLMs), the system achieves reliable self-synchronization of asynchronous multi-view videos, robust target subject identification, and iterative quality control without manual intervention.
We validated the proposed method through synchronous experiments against a Vicon motion capture system.
Quantitative evaluation demonstrated a mean absolute error (MAE) of $5.97^\circ \pm 2.36^\circ$ and a Pearson correlation coefficient of $0.962 \pm 0.014$ in joint angle estimation, indicating high agreement with the gold-standard measurements.
The method showed consistent performance across both healthy subjects and SCI patients, including in noisy and challenging motion patterns.
Overall, the proposed framework offers a practical and reliable solution for long-term kinematic monitoring in neurorehabilitation,
by reducing the burden of patient self-deployment and clinical data processing while ensuring traceable and geometrically grounded joint angle estimation.
The method holds strong potential for daily home-based monitoring and long-term clinical assessment of motor recovery in SCI and related patients.
\section{Acknowledgment}
The authors thank Prof. Guihuai Wang and Dr. Yang Lu for clinical assistant.
\bibliographystyle{unsrt}
\bibliography{References}

\end{document}